\documentclass[10pt,twocolumn,letterpaper]{article}

\usepackage{cvpr}
\usepackage{times}
\usepackage{epsfig}
\usepackage{graphicx}
\usepackage{amsmath}
\usepackage{amssymb}
\usepackage{array,multirow}
\usepackage{booktabs}                  
\usepackage{xcolor}
\newcommand{\rgbbox}[2]{\fcolorbox{black}[rgb]{#1}{\footnotesize \textbf{$\mathtt{#2}$}}}

\usepackage[draft,breaklinks=true,bookmarks=false]{hyperref}

\cvprfinalcopy 

\def\cvprPaperID{****} 

\ifcvprfinal\fi
\begin{document}

\title{Real Time Egocentric Object Segmentation: \\ THU-READ Labeling and Benchmarking Results}

\author{
E. Gonzalez-Sosa, G. Robledo, D. Gonzalez-Morin, P. Perez-Garcia, A. Villegas\\
\textit{Nokia Bell-Labs, Madrid, Spain}\\
\small{\texttt{ester.gonzalez@nokia-bell-labs.com}}
}

\maketitle

\begin{abstract}

Egocentric segmentation has attracted recent interest in the computer vision community due to their potential in Mixed Reality (MR) applications. While most previous works have been focused on segmenting egocentric human body parts (mainly hands), little attention has been given to egocentric objects. Due to the lack of datasets of pixel-wise annotations of egocentric objects, in this paper we contribute with a semantic-wise labeling of a subset of $2124$ images from the RGB-D THU-READ Dataset. We also report benchmarking results using Thundernet, a real-time semantic segmentation network, that could allow future integration with end-to-end MR applications.
\end{abstract}

\vspace{-0.4cm}
\section{Introduction}

In recent years, there has been growing interest in egocentric perception due to the multitude of applications it has in different areas. There are many computer vision tasks related to first person view such as hand and objects detection, activity recognition, visual life-logging, or semantic segmentation, among others. The latter one, is of special relevance for Mixed Reality (MR) applications, as it can be used to introduce elements from the reality into a Virtual Reality (VR) environment, leading to fostering level of presence in VR experiences.   

Previous works have explored computer vision algorithms to integrate user's bodies into VR, either by using color-based approaches \cite{bruder2009enhancing}, depth-based approaches \cite{rauter2019augmenting}, and more recently deep-learning based ones \cite{gonzalez2020enhanced}. One of the main benefits of deep learning approaches compared to color or depth solutions, is that apart from overcoming some of their limitations, they can jointly classify and segment objects of interest. Fulfilling this latter goal demands several challenges to be solved. First, to train deep learning networks for both segmentation and classification tasks, it is required a dataset with pixel wise semantic labels. Second, for MR applications, networks need to perform in real time, constraining their maximum inference times.

In this work, we introduce a new dataset for egocentric object segmentation together with pixel-wise labeling, which will be made publicly available for the research community upon request\footnote{URL will be added after acceptance of the paper.}. In addition, a benchmark is provided which serves as starting point for developing new egocentric object segmentation methods. As we are particularly interested on integrating this network in MR applications such that users can interact with real objects while wearing the VR googles, we only focus on real time semantic segmentation networks. 

\begin{figure*}
\centering
\includegraphics[width=1.0\linewidth]{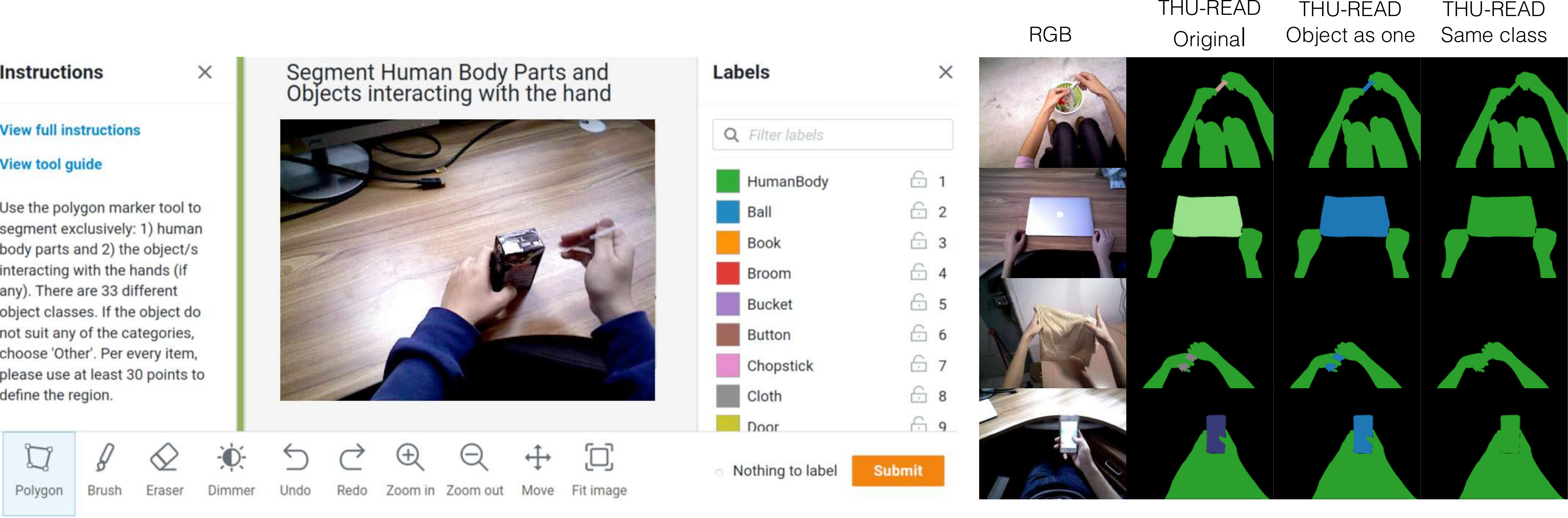} 
\caption{Left: Designed Segmentation Annotation Pipeline; Right: Example of the different THU-READ groundtruth versions.}\vspace{-0.3cm}
\label{dataset}
\end{figure*}

The remainder of this paper is structured as follows: Section~\ref{related_works} summarises earlier works focused on egocentric segmentation. Section~\ref{egobody_dataset} motivates the decision for THU-READ dataset and describes the procedure conducted to obtain a pixel-wise labeling from it. Section~\ref{results} explains the protocol and reports the results obtained for the different versions of the THU-READ groundtruth. Finally, Section~\ref{conclusion} concludes the paper with some discussions and future research lines.

\section{Related Works}
\label{related_works}

This section briefly reviews previous research attempts to segment elements (either human body or objects) from egocentric views, covering mainly works that meet real time requirements, and thus, have been or could be integrated in MR applications.  

In this regard, the simplest methods are those based on color-based approaches \cite{fiore2012towards,bruder2009enhancing,immersirve_gastronomic2019,gunther2015aughanded}, which can be deployed in real time but tend to work well just under controlled conditions. Besides, the segmented objects are constrained by color appearance. Contrarily, depth solutions are based on the incorporation of real objects placed within a distance from the camera \cite{rauter2019augmenting}. Thus, depth could be the solution for applications in which desired objects are all within that range.

While color-based and depth-based solutions manage to segment objects under certain constrains, they failed at classifying those objects. To achieve this, one has to move towards deep learning approaches, and more in particular to semantic segmentation networks. In this line, Gonzalez-Sosa \textit{et al.} explore state-of-the-art networks for segmenting egocentric human body parts \cite{gonzalez2020enhanced} using a semi-synthetic dataset. The reported results suggest suggested that semantic segmentation networks can overcome some of the limitations of color-based solutions, as now networks incorporate more complex information than just color, or of depth-based solutions, specially in outdoors scenarios.

\section{THU-READ Segmentation Labeling}
\label{egobody_dataset}

Although in this work the segmentation approach will be purely based on RGB images, to maximize the impact of the labeling effort, we prioritize the search of datasets that contain egocentric images from both RGB and depth modalities. To the best of our knowledge, there exist only three published egocentric and RGB-D datasets including human body parts (and) objects: Ego Gesture \cite{zhang2018egogesture}, created for the task of hand gesture recognition, THU-READ \cite{tang2018multi}, developed for the task of action recognition and First-Person Hand Action Benchmarch (FPHA) with RGB-D \cite{garcia2018first}, also designed to recognize first-person dynamics interacting with 3D objects. None of them is providing pixel-wise segmentation labels. Among these three, we are interested to get pixel-wise labeling from THU-READ frames because, unlike EgoGesture, its depth frames show continuous distance values, which will be more valuable to infer the geometry information of the egocentric view. We also discarded FPHA since the mo-cap system is placed on all hand dorsum’s. 

THU-READ is a RGB-D dataset collected at Tsinghua University, designed for recognizing egocentric actions which have some relationship with hands \cite{tang2018multi}. It contains recordings of $40$ different actions from $8$ different subjects, repeated $3$ times, making a total of $960$ RGB-D videos. 

In this work a representative subset of $640\times480$ images from all users, actions and repetitions is created through video sampling. Fig.\ref{dataset} depicts the labeling tool selected from the semantic segmentation template of AMT\footnote{\url{https://requester.mturk.com/create/projects/new}}. In our case, the Human Intelligence Task (HIT) consisted on using the polygon marker tool to define the boundaries of $1)$ human body parts and, $2)$ objects interacting with them. From the $40$ different egocentric actions defined in THU-READ, we define objects categories to account for objects which were interacting with the hands, or that do have a lot to do with the action being made\footnote{Notice that there were videos with more objects than those requested to be labeled, or there were actions where no object was involved, e.g \textit{thumbs up.}}, e.g: \textit{open drawer}. The following $30$ different object categories were obtained: ball, book, broom, dustpan, cutlery, cloth/towel, cup/pot, drawer, fruit, glove, knob, laptop, mouse, nail clipper, paper, paper-plane, pen, phone, plug, stapler, toothbrush, toothpaste, straw, umbrella, watch, watering can, weight, case/backpack, scissors and other. 
\begin{table*}[]
    \centering
    \caption{IoU for the different THU-READ Groundtruth versions. IoU is reported in the range $0-100\%$.}
    \tiny
    \label{tab:results}
    \resizebox{.5\textwidth}{!}{
\renewcommand*{\arraystretch}{1.2}
\begin{tabular}{llllll}
\toprule
   \textbf{Dataset}  & \textbf{IoU Background} & \textbf{IoU Human Body}  & \textbf{IoU mean Objects}    & \textbf{IoU mean} & \textbf{Inference Time}      \\ \midrule
THU-READ original & $94.00$ & $76.00$ & $28.57$ & $66.19$ & $15ms$ \\ 
THU-READ objects as one & $94.00$ & $77.70$ & $42.40$ & $71.36$ & $15ms$   \\ 
THU-READ same label& $92.80$ & $69.40$ & $--$ & $81.10$ & $15ms$   \\ 
\bottomrule
\label{IoU_results}
\end{tabular}
}\vspace{-0.4cm}
\end{table*}

\begin{table}[]
    \centering
    \caption{IoU for the different $30$ object categories of the network trained with THU-READ original. IoU is reported in the range $0-1$.}
    \label{tab:results_2}
\resizebox{.495\textwidth}{!}{
\renewcommand*{\arraystretch}{1.2}
\begin{tabular}{lllllllll}
\toprule
   \textbf{RGB} & \textbf{Class}  & \textbf{IoU} &  \textbf{RGB} & \textbf{Class}  & \textbf{IoU} & \textbf{RGB} & \textbf{Class}  & \textbf{IoU}     \\ \midrule
\rgbbox{0,0,0}{\color{white}000080} & Background	&	$0.94$	&	\rgbbox{0.682,0.780,0.909}{\color{white}AEC7E8} & Glove	&	$0.86$	&	\rgbbox{0.6110,0.619,0.870}{\color{white}9C9EDE} & Stapler	&	$0.19$	\\
\rgbbox{0.172,0.627,0.172}{2CA02C} & Human body	&	$0.76$	&	\rgbbox{1.0,0.733,0.470}{\color{white}FFBB78} & Knob	&	$0.38$	&	\rgbbox{0.388,0.474,0.105}{\color{white}637939} & Toothbrush	& $0.12$	\\
\rgbbox{0.121,0.466,0.705}{1F77B4} & Ball	&	$0.39$	&	\rgbbox{0.596,0.874,0.541}{\color{white}98DF8A} & Laptop	&	$0.14$	&	\rgbbox{0.549,0.635,0.321}{\color{white}8CA252} &Toothpaste	&	$0.13$	\\
\rgbbox{1.0,0.498,0.054}{FF7F0E} & Book	&	$0.09$	&	\rgbbox{0.772,0.690,0.835}{\color{white}C5B0D5} & Mouse	&	$0.18$	&	\rgbbox{0.709,0.811,0.419}{\color{white}B5CF6B} & Straw	&	$0.00$	\\
\rgbbox{0.839,0.152,0.156}{D62728} & Broom	&	$0.15$	&	\rgbbox{0.768,0.611,0.580}{\color{white}C49C94} & Nail Clipper	&	$0.00$	&	\rgbbox{0.807,0.858,0.611}{\color{white}CEDB9C} & Umbrella	&	$0.58$	\\
\rgbbox{0.580,0.403,0.741}{9467BD} & Dustpan	&	$0.12$	&	\rgbbox{0.968,0.713,0.823}{\color{white}F7B6D2} & Other	&	$0.18$	&	\rgbbox{0.549,0.427,0.192}{\color{white}8C6D31} & Watch	&	$0.12$	\\
\rgbbox{0.890,0.466,0.760}{E377C2} & Cutlery	&	$0.10$	&	\rgbbox{0.780,0.780,0.780}{\color{white}C7C7C7} & Paper	&	$0.50$	&	\rgbbox{0.741,0.619,0.223}{\color{white}BD9E39} & Watering Can	& $0.81$	\\
\rgbbox{0.498,0.498,0.498}{7F7F7F} & Cloth/Towel	&	$0.17$	&	\rgbbox{0.858,0.858,0.552}{\color{white}DBDB8D} & Paperplane	&	$0.77$	&	\rgbbox{0.905,0.729,0.321}{\color{white}E7BA52} & Weight	&	$0.40$	\\ 
\rgbbox{0.737,0.741,0.133}{BCBD22} & Cup/Pot	&	$0.09$	& \rgbbox{0.619,0.854,0.898}{\color{white}9EDAE5} &	Pen	&	$0.03$	&	\rgbbox{0.905,0.796,0.580}{\color{white}E7CB94} & Case/Backpack	&	$0.83$	\\
\rgbbox{1.0,0.596,0.588}{FF9896} & Drawer	&	$0.17$	&	\rgbbox{0.223,0.231,0.474}{\color{white}393B79} & Phone	&	$0.73$	&	\rgbbox{0.223,0.235,0.517}{\color{white}000080} & Scissors	&	$0.02$	\\
\rgbbox{0.090,0.745,0.811}{17BECF} & Fruit	&	$0.32$	&	\rgbbox{0.419,0.431,0.811}{\color{white}6B6ECF} & Plug	&	$0.00$	& -- &	mean IoU	&	$0.28$	\\

\bottomrule
\label{IoU_results}
\end{tabular}
}\vspace{-0.4cm}
\end{table}

As part of the instructions, a segmentation guideline to assure that the HIT requirements were properly understood\footnote{\url{https://bit.ly/33jXegd}} was included, for instance: a minimum of $30$ points per boundary were required to consider the labeling process successful. Turks were not paid automatically after finishing their task. The authors had a period of some days to check the quality of the segmentation and labeling. 

From this AMT-based labeling experiment we found of special relevance that: $i)$ better results were obtained from Turks who hold Master Qualification; and $ii)$ instructors checked themselves that boundaries were precise enough and classes were corrected\footnote{Done by overlapping labeled images created by Turks on top of their original RGB counterpart images}. If images were not correctly processed, instructors rejected the tasks, providing detailed feedback. Once labeled images were accepted, Turks received a compensation in the range of $25-40$ cents per HIT, comprising also Amazon Fee and 5\% extra for Master Qualification\footnote{Labeling was sponsored by Amazon and was created for the EPIC @ CVPR 2020 Dataset challenge \url{https://eyewear-computing.org/EPIC_CVPR20/mturk-proposals}}.
As a result, we obtained a $2124$ subset of labeled images.
For a more insightful benchmarking process, different groundtruth version were created (see examples of Fig.\ref{dataset}):

\begin{itemize}
\item \textbf{THU-READ original}: composed of $32$ categories including human body, background, and the $30$ object categories already mentioned.\vspace{-0.2cm}
\item \textbf{THU-READ objects as one}: human body, background and objects ($3$ classes). All different objects were assigned to the same class.\vspace{-0.2cm}
\item \textbf{THU-READ same class}: human body and objects as a single class ($2$ classes).\vspace{-0.2cm}
\end{itemize}

The motivation for creating also THU-READ objects as one, and THU-READ same class was two-fold: $i)$ to better understand the impact of number of classes and class intra-variability on the performance; and $ii)$ to study the benefits of a simplified case where objects of interest are those interacting with the hands, regardless of their particular class.  

\section{Experimental Protocol and Results}
\label{results}

As our aim with egocentric object segmentation is planned to be part of an end-to-end MR application, a strong real time inference time is required. As a result, benchmarking experiments were conducted using a real-time semantic segmentation network from our previous work \cite{gajic2020egocentric}, based on the Thundernet network proposed by Xiang \textit{et al.} \cite{xiang2019thundernet} that we modified for larger and egocentric images.

Thundernet was developed and trained using Keras framework version 2.2.4, Python 3.5, and used with two GPU GTX-1080 Ti with 12GB RAM each. The weights from the three Resnet-18 blocks inside the encoder were inherited from a model pre-trained on ImageNet dataset. Then, the whole architecture was fine-tuned with the corresponding THU-READ subset version. To assure that validation and training sets were disjoint, all images pertaining to one of the $8$ users were taken for validation ($302$) whereas the remainder ($1822$) were used for training. Results are reported in terms of Intersection over Union (IoU). 

An exhaustive set of experiments following grid search strategies varying number of epochs, learning rate and weight decay were conducted, monitoring validation performance.The final training configuration was: learning rate of $1e-4$, $20$ epochs, weight decay of $2e-4$, and batch size of $4$ for THU-READ original; learning rate of $1e-4$, $20$ epochs, weight decay of $2e-5$, and batch size of $8$ for THU-READ objects as one; and learning rate of $1e-4$, $20$ epochs, no weight decay, and batch size of $4$ for THU-READ same label. In all cases Adam Optimizer was used.

\begin{figure*}
\centering
\includegraphics[width=0.9\linewidth]{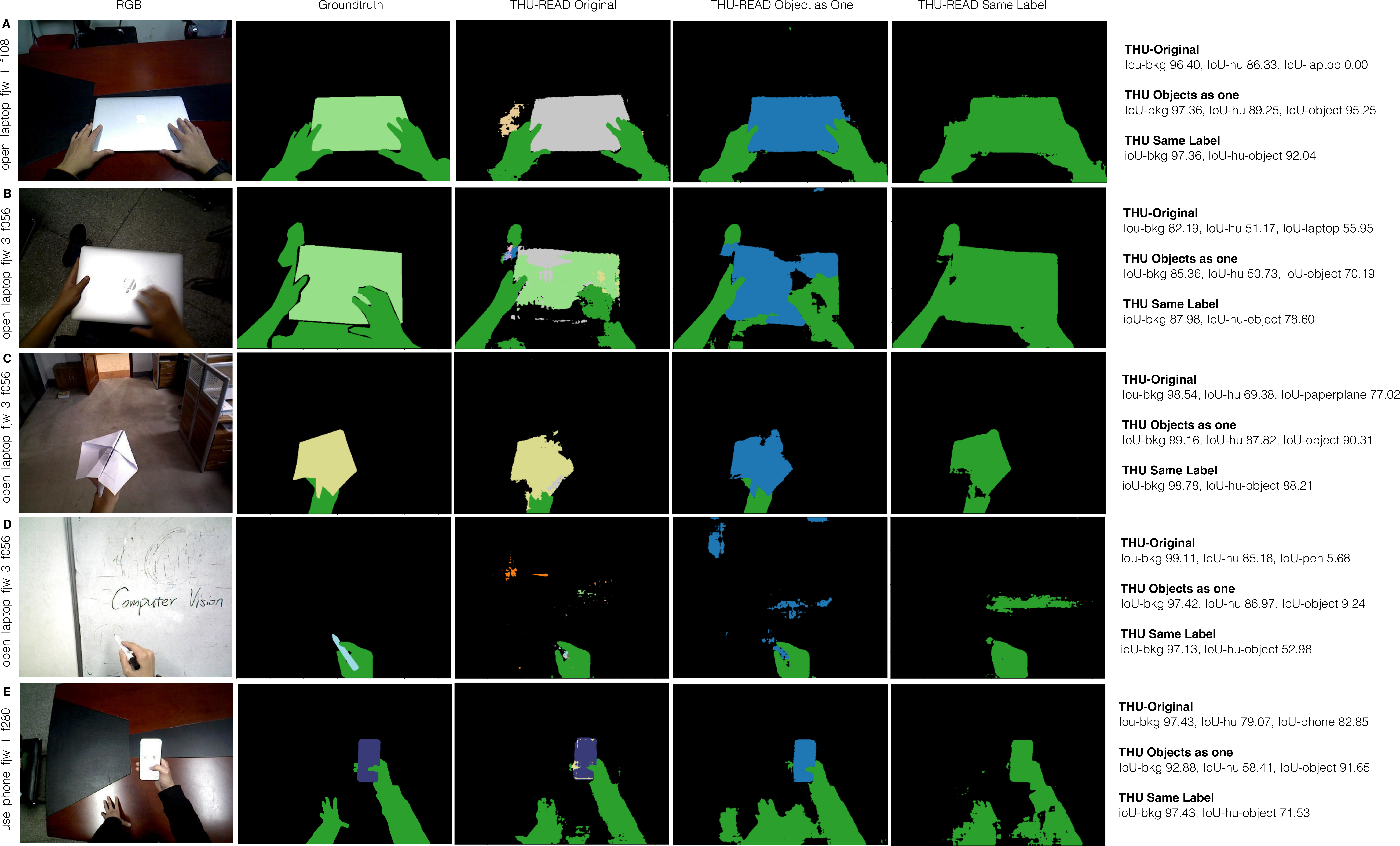} 
\caption{Several examples of egocentric objects and human body parts being segmented using Thundernet trained with different THU-READ groundtruth versions.}
\label{example_images}\vspace{-0.4cm}
\end{figure*}

As can be seen from Table \ref{tab:results}, while similar performances are achieved for segmenting human bodies for both THU-READ original and THU-READ object as one, there is a significant performance increase from $28.57\%$ to $42.40\%$ IoU for the object categories. One justification could be that in the second case, the network is segmenting the objects relying more on spatial information, objects in the hand area, rather than on their particular appearance or shape. No significant difference was found between THU-READ objects as one and THU-READ same label, although at qualitative level one can see examples whether segmentation accuracy is better (see Fig.\ref{example_images} B). 

If we have a closer look at the individual IoU reported in Table \ref{tab:results_2} for the THU-READ Original case, it can be observed that high IoU values are reported for either large objects, or for objects which are easily distinguishable from the background (see Fig.\ref{example_images} C). It also seems that the network fails at segmenting small objects such as pen or straw, possibly due to their size partial occlusions. Further analysis are needed to understand in which conditions there are problem of poor segmentation quality or confusion between classes (see Fig.\ref{example_images} A.). Besides, THU-READ images vary a lot in terms of illumination, and scenarios, so more in-depth  analysis and more datasets are required to extract more reliable conclusions.


\section{Conclusions}
\label{conclusion}
In order to facilitate the development of egocentric object segmentation we introduced a new pixel-wise labeled dataset which is made publicly available to the research community. Additionally, we provided a benchmark which may serve as baseline for future developments in the field. 

Various avenues for future work can be considered, e.g. the exploration of depth as an additional source, and considering more datasets for a better problem understanding. 


{\small
\bibliographystyle{ieee_fullname}
\bibliography{egbib}
}

\end{document}